\documentclass[runningheads]{llncs}

\usepackage{eccv}

\usepackage{eccvabbrv}

\usepackage{graphicx}
\usepackage{booktabs}
\usepackage{multirow}
\usepackage{tabularx}
\usepackage{graphicx}
\UseRawInputEncoding

\usepackage[accsupp]{axessibility}  

\usepackage{hyperref}

\usepackage{orcidlink}

\begin{document}

\title{Reasoning3D - Grounding and Reasoning in 3D: Fine-Grained Zero-Shot Open-Vocabulary 3D Reasoning Part Segmentation via Large Vision-Language Models} 

\titlerunning{Reasoning3D - Grounding and Reasoning in 3D}

\author{Tianrun Chen\inst{1,2}$^{+}$ \and
Chunan Yu\inst{3}$^{+}$, Jing Li \inst{3}, Jianqi Zhang \inst{3}, Lanyun Zhu\inst{4}, Deyi Ji\inst{6}, Yong Zhang \inst{3}, Ying Zang  \inst{3}$^{*}$, Zejian Li\inst{5}, Lingyun Sun\inst{2}}

\authorrunning{T, Chen et al.}

\institute{\href{http://www.kokoni3d.com/}{KOKONI, Moxin (Huzhou) Tech. Co., LTD.} \and
College of Computer Science and Technology, Zhejiang University. \and
School of Information Engineering, Huzhou University. \and 
Information Systems Technology
and Design Pillar, Singapore University of Technology and Design.\and
School of Software Technology, Zhejiang University. \and
School of Information Science and Technology, University of Science and Technology of China.\\ 
+ Equal Contribution    * Corresponding Author \\
\email{tianrun.chen@zju.edu.cn; 02750@zjhu.edu.cn}\\
Project Page: \href{http://tianrun-chen.github.io/Reason3D/}{http://tianrun-chen.github.io/Reason3D/}
}

\maketitle

\begin{figure*}[h]
\centering
\includegraphics[width=0.76\textwidth]{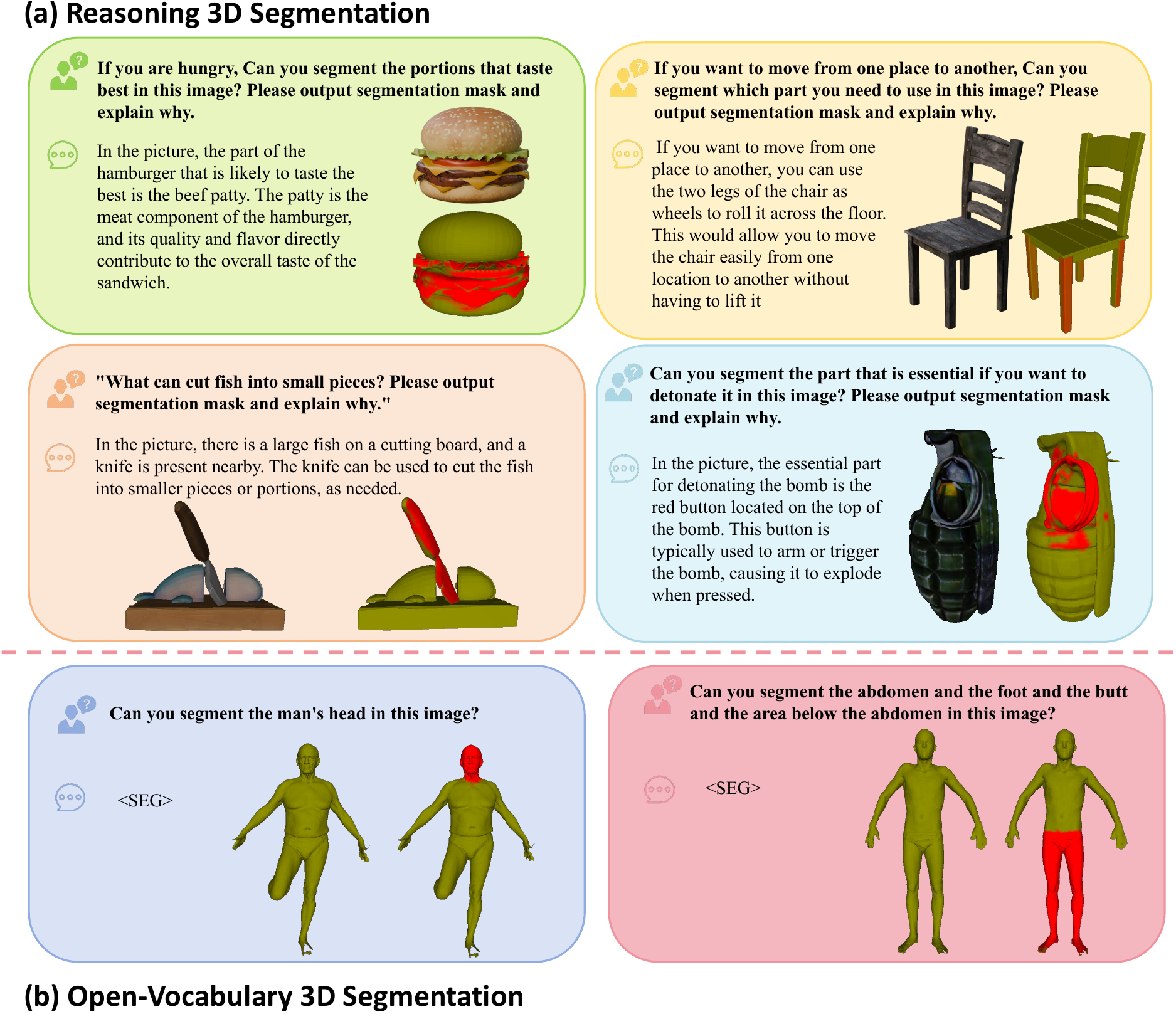}
\caption{In this work, we propose a new task: reasoning 3D segmentation. We also propose a method that can segment 3D object parts with explanations based on various criteria such as reasoning, shape, location, function, and conceptual instructions.} \label{fig1}
\label{sketchin}
\end{figure*}

\begin{abstract}
  In this paper, we introduce a new task: Zero-Shot 3D Reasoning Segmentation for parts searching and localization for objects, which is a new paradigm to 3D segmentation that transcends limitations for previous category-specific 3D semantic segmentation, 3D instance segmentation, and open-vocabulary 3D segmentation. We design a simple baseline method, Reasoning3D, with the capability to understand and execute complex commands for (fine-grained) segmenting specific parts for 3D meshes with contextual awareness and reasoned answers for interactive segmentation. Specifically, Reasoning3D leverages an off-the-shelf pre-trained 2D segmentation network, powered by Large Language Models (LLMs), to interpret user input queries in a zero-shot manner. Previous research have shown that extensive pre-training endows foundation models with prior world knowledge, enabling them to comprehend complex commands, a capability we can harness to "segment anything" in 3D with limited 3D datasets (source efficient). Experimentation reveals that our approach is generalizable and can effectively localize and highlight parts of 3D objects (in 3D mesh) based on implicit textual queries, including these articulated 3d objects and real-world scanned data. Our method can also generate natural language explanations corresponding to these 3D models and the decomposition. Moreover, our training-free approach allows rapid deployment and serves as a viable universal baseline for future research of part-level 3d (semantic) object understanding in various fields including robotics, object manipulation, part assembly, autonomous driving applications, augment reality and virtual reality (AR/VR), and medical applications. The code, the model weight, the deployment guide, and the evaluation protocol are: \href{http://tianrun-chen.github.io/Reason3D/}{http://tianrun-chen.github.io/Reason3D/}.
\end{abstract}

\section{Introduction}
\label{sec:intro}
The importance of 3D segmentation cannot be overstated \-- it is foundational in fields like robotics, autonomous driving, and augmented reality\cite{PartNet, SHRED, ZeroPS}. Traditional approaches have often required extensive manual labeling or complex rule-based algorithms that struggle to generalize to diverse real-world scenarios\cite{HAL3D, OmniSeg3D, PartSLIP, lyu2020learning}. The sheer complexity of 3D data, combined with the inherent ambiguities and varying viewpoints, has posed significant challenges in developing robust and generalizable 3D segmentation techniques.  

In this work, we introduce fine-grained \textbf{Zero-Shot 3D Reasoning Segmentation} for parts in 3D objects, which aims to bring 3D segmentation to a new level. Imagine instructing a system with words like "segment the part of the chair where you would sit" or "highlight the nutritious parts of this vegetable" and watching it magically understand and perform the task in the 3D world. You can have natural conversations with the system and see it output the segmentation mask along with explanations (See Fig. \ref{fig1} for examples). It is a future where 3D systems can intuitively understand and respond to intricate queries -- The possibilities are endless.

However, achieving this vision is no small feat. Traditional 3D segmentation approaches are typically confined to fixed object categories, severely limiting their flexibility. Recent endeavors in open-vocabulary segmentation can handle a broader range of labels but are still limited to dealing with straightforward tasks like labeling "the apple" and cannot handle complex, reasoning-based queries. Asking a system to perform more nuanced operations like "segment edible parts of a fruit" requires a level of contextual understanding and reasoning that current methods do not possess. 

Thanks to the recent advancements in Multi-modal Large Language Models (LLMs)\cite{InternGPT, HuggingGPT, M610T, CogVLM, MiniGPT5, zhu2024ibd}, we can now bring our aforementioned vision to life. Recently, Large Vision-Language Models (LVLM) have shown remarkable capabilities in comprehending 2D images, excelling in tasks that require complex reasoning, multi-turn conversations, and explanatory answers\cite{LISA, LISA++, LLaFS}. We aim to extend their capabilities into the 3D realm, and we believe that this transition is promising with much practical value -- never forget that we live in a 3D world!

However, extending the success of reasoning segmentation from 2D to 3D domains also presents substantial challenges. The scarcity of available 3D data and ground-truth Question-and-answer pairs stopped us from performing large-scale training. The added dimension also increases the computational demands of 3D architectural components. Here, inspired by research that has tackled similar challenges in 3D generation\cite{Xiang_2023_ICCV, MVDream, Make-It-3D, Panoptic_NeRF, PanopticNeRF_360, zhang2023painting} -- using network models in 2D and then lifting some information to 3D, we introduce our approach to leverage off-the-shelf 2D models to perform the task in a zero-manner. This approach, which we named Reasoning3D, allows us to circumvent the limitations imposed by the scarcity of extensive 3D datasets and the high computational costs with its training-free property and 2D pre-training. 

Specifically, our Reasoning3D approach involves rendering a 3D model from multiple viewpoints and applying a pre-trained reasoning segmentation network to each 2D view based on the given query input. By doing so, we generate segmentation masks and accompanying text explanations for each perspective. These individual masks and explanations are then fused to produce a comprehensive 3D segmentation mask (labels are assigned to the vertices of the 3D model). We have evaluated our approach in various models in the wild, both with and without textures. We have also tested our approach in existing open-vocabulary segmentation benchmarks, which validates the effectiveness of our approach. 

While Reasoning3D is a straightforward baseline method, we believe it serves as a good starting point for researchers to explore and expand the future of 3D part segmentation. We will release the implementation code and the benchmark code publicly to facilitate future research, with the hope that our initial step sets the stage for further innovation and refinement, and eventually bring us closer to a future where 3D computer vision systems are as versatile and perceptive as human cognition, capable of revolutionizing a myriad of applications across various fields. 

\section{Related Work}

\subsection{3D Semantic Segmentation.}
Segmentation in 2D scenes has achieved significant progress in recent years \cite{zhu2021learning, zhu2023continual, wang2023fvp, zang2024resmatch, chen2023sam}, yet understanding and reasoning in 3D environments is still a crucial research area that needs more attention. In the domain of 3D semantic segmentation, our objective is to predict the semantics of each point in a point cloud. Notable advancements in this field encompass point-based approaches \cite{Huang_Wang_Neumann_2018, Charles_Su_Kaichun_Guibas_2017}, leveraging sophisticated point convolution techniques \cite{Thomas_Qi_Deschaud_Marcotegui_Goulette_Guibas_2019, Xu_Ding_Zhao_Qi_2021}, and voxel-based approaches \cite{Choy_Gwak_Savarese_2019, Graham_Engelcke_Maaten_2018}. Some techniques generate point-level segmentation results using 3D sparse convolutions [11], while others utilize transformer-based approaches \cite{Lai_Liu_Jiang_Wang_Zhao_Liu_Qi_Jia}. Furthermore, multi-view semantic segmentation methods such as DeepViewAgg \cite{Robert_Vallet_Landrieu_2022}, Diffuser \cite{Kundu_Yin_Fathi_Ross_Brewington_Funkhouser_Pantofaru_2020, Mascaro_Teixeira_Chli_2021}, 3D-CG \cite{Hong_Du_Lin_Tenenbaum_Gan_2022}, and 3D-CLR\cite{Hong_Lin_Du_Chen_Tenenbaum_Gan_Ucla} enhance representation learning by creating 2D projections of 3D scenes from different viewpoints. Studies have demonstrated that multi-view representations effectively improve the performance and robustness of various 3D tasks. Nevertheless, these methods usually depend on predefined semantic label sets, whereas our approach is tailored to address and interpret complex reasoning queries. We believe that, following the trend that researchers trying to use various inputs to improve the computer-human interaction in 3D models\cite{chen2023deep3dsketch,chen2023reality3dsketch,chen2024deep3dsketchim,zang2023deep3dsketch+,zang2023deep3dsketch+2,chen2023deep3dsketch+}, this work can provide another avenue for manipulating 3D contents.

\subsection{Large Multimodal Models.} Extensive research on large language models (LLMs) has highlighted their reasoning capabilities, leading to efforts to expand these abilities into the visual domain using large multimodal models (LMMs). LMMs are highly adaptable and versatile, capable of performing tasks that require both language and vision skills. Notable models like BLIP-2 \cite{Li_Li_Savarese_Hoi}, LLaVA \cite{Liu_Li_Wu_Lee_}, and MiniGPT-4\cite{Zhu_Chen_Shen_Li_Elhoseiny} typically utilize a two-phase training approach, which aligns visual representations with LLMs' linguistic embeddings through extensive image-text and video-text datasets \cite{Bain_Nagrani_Varol_Zisserman_2021, Changpinyo_Sharma_Ding_Soricut_2021, Lin_Maire_Belongie_Hays_Perona_Ramanan_Dollár_Zitnick_2014, Miech_Zhukov_Alayrac_Tapaswi_Laptev_Sivic_2019, Schuhmann_Beaumont, Schuhmann_Vencu_Beaumont, Sharma_Ding_Goodman_Soricut_2018}. Recently, the focus has been on merging multimodal LLMs with vision tasks. VisionLLM \cite{Wang_Chen_Chen_Wu_Zhu_Zeng_Luo_Lu_Zhou_Qiao_et}, for example, offers a versatile interface for various vision-centric tasks via instruction tuning. Nevertheless, this model does not fully leverage the sophisticated reasoning capabilities of LLMs. Kosmos-2 \cite{Peng_Wang_Dong_Hao_Huang_Ma} seeks to bolster the foundational abilities of LLMs by creating large datasets of aligned image-text pairs. DetGPT \cite{Pi_Gao_Diao_Pan_Dong_Zhang_Yao_Han_Xu_Kong_et} smoothly combines a fixed multimodal LLM framework with an open-vocabulary detector to enable instruction-based detection. LISA, LISA++ \cite{LISA, LISA++} generates segmentation masks using embeddings from vision-language models, and LLaFS generates segmentation masks using coordinates exported from LLM. GPT4RoI \cite{Zhang_Sun_Chen_Xiao_Shao_Zhang_Chen_Luo} presents an innovative method by incorporating spatial boxes as inputs and training on region-text pairs. 

Unlike previous approaches, our approach aims to integrate the vision-language capabilities of LMMs with the reasoning strengths of LLMs in new 3D perception tasks, taking advantage of these developments in the LMM field.

\subsection{Language Instructed 3D Tasks.} The fusion of point clouds with natural language processing has profound implications, generating significant interest in the field of 3D scene comprehension. This rapidly evolving domain holds promises for advancing human-robot interaction, metaverse development, robotics, and embodied intelligence. Two pivotal abilities crucial to 3D environment dialogue systems include spatial perception and logical reasoning.
Recently, there has been a surge in tasks integrating 3D scenes and languages, such as 3D captioning, question answering, situated Q and A, embodied dialogue, planning, navigation, multi-turn dialogue assistance, object detection, and scene description. We categorize 3D perception task models into three groups (refer to Table 1, split by dashed lines). The first group encompasses models handling tasks like 3D captioning, situated question answering, and visual grounding \cite{Ma_Yong_Zheng_Li_Liang_Zhu_Huang_2022, Zhu_Ma_Chen_Deng_Huang_Li_2023}. These models can generate single words or phrases as textual outputs. The second category consists of 3D semantic segmentation models producing 3D segmentation masks, such as 3DOVS \cite{Liu_Zhan_Zhang_Xu_Yu_Saddik_Theobalt_Xing_Lu_2023}, Openmask3D \cite{Takmaz_Fedele_Sumner_Pollefeys_Tombari_Engelmann_2023}, OpenScene \cite{Peng_Genova_Jiang_Tagliasacchi_Pollefeys_Funkhouser_2022}, and PLA \cite{Ding_Yang_Xue_Zhang_Bai_Qi_2022}, which perform open-vocabulary semantic segmentation for 3D scenes. However, these methods do not offer conversational responses to user queries or provide reasoning for their tasks. The third category comprises models employing LLMs to conduct visual perception tasks like captioning, scene understanding, and visual grounding, offering conversational outputs \cite{chen2023ll3da, Guo_Zhang_Zhu_Tang_Ma_Han_Chen_Gao_Li_Li_et, Hong_Zhen_Chen_Zheng_Du_Chen_Gan, li2023m3dbench, wang2023chat3d, xu2023pointllm, Yang_Chen_Qian_Madaan_Iyengar_Fouhey_Chai_2023}. Nonetheless, they lack fine-grained semantic segmentation or reasoning-based 3D vision tasks.

\section{Method}
\begin{figure*}[htb]
\centering
\includegraphics[width=1.0\textwidth]{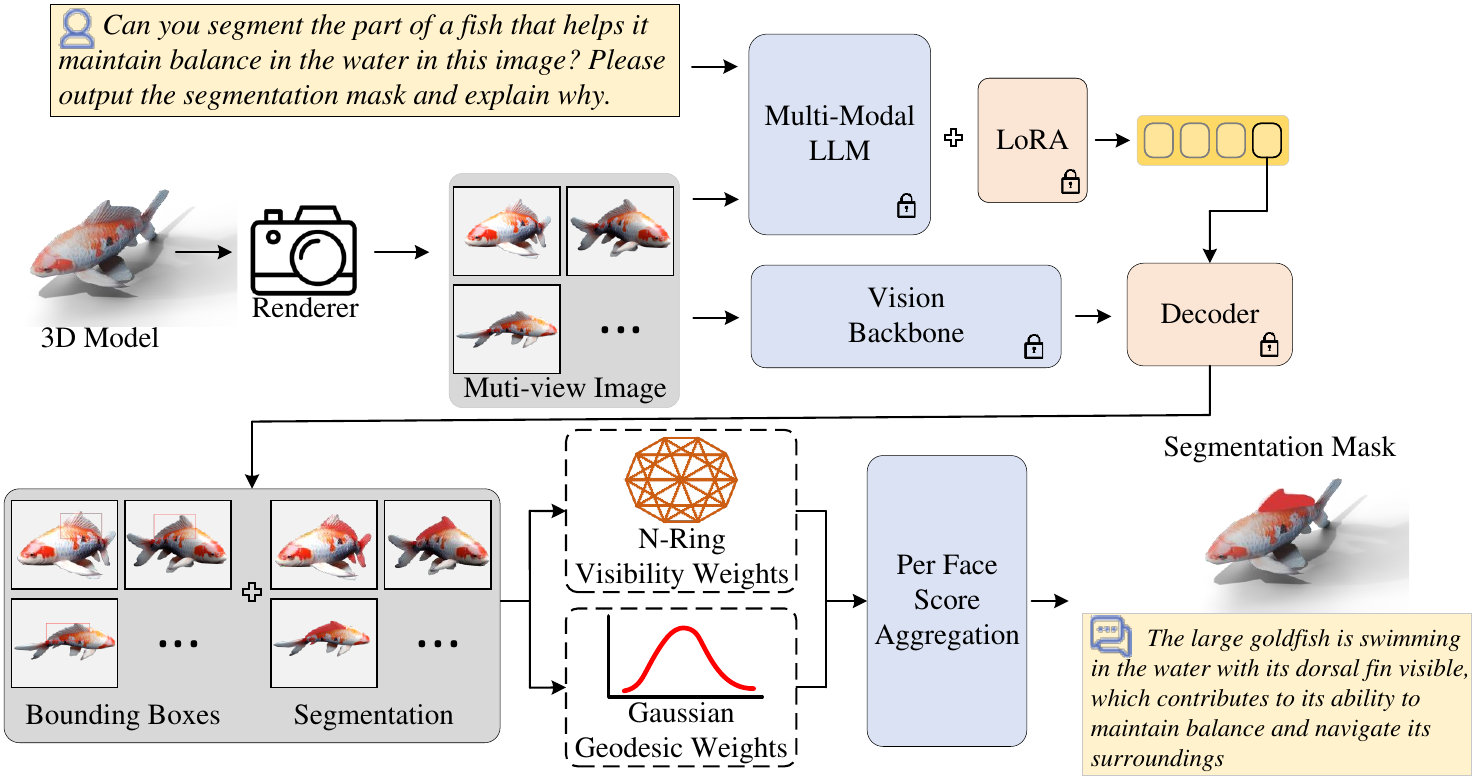}
\caption{The overview of Reasoning3D. First, a 3D model represented by 3D meshes is fed into a renderer to obtain multi-view images. Then, each image goes through a vision backbone and a multi-modal LLM along with user input queries. The decoder decodes the final layer embedding which contains the extra token, thus producing K segmentation masks. We also extract the bounding boxes in this stage. Finally, a specially designed mask-to-3D segmentation algorithm elevates the projections back into the 3D space.} \label{Fig2}
\end{figure*}

As illustrated in Fig. \ref{Fig2}, Reasoning3D begins with a mesh input fed into the renderer for viewpoint rendering, generating the face ID for each corresponding viewpoint. Next, the rendered viewpoints and the user-input prompt are processed by the pre-trained 2D reasoning segmentation network, which segments the image to extract the desired parts and output explanations. Finally, using the mapping relationship between each viewpoint and its corresponding mesh face ID, the segmented parts are reconstructed back onto the mesh with a specially designed multi-view fusion mechanism.

\subsection{Multi-View Image Rendering and Face ID Generation.}
Human interaction with the 3D environment often involves dynamic exploration, amalgamating viewpoints from various angles to construct a cohesive 3D comprehension, rather than assimilating a 3D setting instantaneously. Our methodology advocates for 3D reasoning cultivated from multi-perspective imagery. This strategy also leverages the extensive 2D pretraining accessible in vision-language models, akin to prior methodologies capitalizing on pre-trained vision-language models for 3D visual tasks. The input for this process is mesh $F=\{f_n\}_{n=1}^{N}$, which is composed of N sets of faces $f_n$. During this rendering process, the 3D model is converted into multiple 2D images $X_{img}=\{x_i\}_{i=1}^{11}$ from various perspectives. In addition to generating these 2D images, the rendering process also produces corresponding face IDs for each image. These face IDs serve as a crucial link between the 2D images and the original 3D mesh. Specifically, they form a mapping matrix $W_{pf}$ that connects each pixel $P=\{p_i\}_{i=k}^{Mm}$ in the 2D images to a specific face $f_n$ on the 3D mesh, ensuring ensures that the 2D and 3D data remain accurately aligned. The process is formulated as follows:
\begin{align}
    \begin{split}
        \mathit P = W_{pf}\sum_{n=1}^{N}f_{n}
    \end{split}
\end{align}

where $P$ represents the pixels in the rendered images, and $f_n$ represents the faces of the 3D mesh from the viewpoint. Denoted by $k$, the face ID $W_{pf}$ connects the pixels $P$ in the rendered image and the faces of the 3D mesh $f_n$ visible in the current view. 

\subsection{Reasoning and Segmenting with User Input Prompt}
Unlike previous methods (e.g. CLIPSeg\cite{CLIPSeg}, LSeg\cite{LSeg}, and GLIP\cite{GLIP}) which can handle open-vocabulary explicit prompt, our method aims to handle the implicit prompt such as "Can you segment the appropriate parts of the image containing a 'caged bird'?" Here, we leverage the recent advances of large foundation models to perform this multi-modal reasoning task. 

Following Lai et al.,\cite{LISA}, we extend the original LLM vocabulary with a new token, <SEG>, which denotes a request for segmentation output. Given the user-input prompt question $X_{question}$ and the input image $X_{img}$, these are input into the multimodal large language model (LLM)$F_{MM}$, which subsequently outputs the textual response $Y_{answer}$. The process is formulated as:

\begin{align}
    \begin{split}
        \mathit Y_{answer} = F_{MM}(X_{img}, X_{question})
    \end{split}
\end{align}

Next, the generation of segmentation masks corresponding to the input images involves a series of steps. Firstly, we extract the embedding $\hat{E}_{answer}$ corresponding to the <SEG> token from the output answer textual $Y_{answer}$. This step enables us to capture information relevant to the segmentation task from the language prompts. Subsequently, we process $\hat{E}_{answer}$ through the MLP $\gamma$ projection layer to obtain the feature vector $E_{answer}$. Concurrently, utilizing a visual backbone network $F_{vb}$, we extract visual embeddings $E_{img}$ from the visual inputs $X_{img}$. Finally, we feed both the feature vector $E_{answer}$ and the visual embeddings $E_{img}$ into the decoder $F_{dec}$. The decoder $F_{dec}$ utilizes these features to generate the final segmentation mask $M$ and confidence scores $S_{M}$ for each mask. This yields segmentation results based on both language prompts and visual information, where each segmentation mask is accompanied by its respective confidence score and corresponding answer textual. The detailed structure of the decoder follows Segment Anything \cite{SAM}. The process is formulated as follows:

\begin{align}
    \begin{split}
        \mathit E_{answer} = \gamma \hat{E}_{answer}
    \end{split}
\end{align}

\begin{align}
    \begin{split}
        \mathit E_{img} = F_{vb}(X_{img})
    \end{split}
\end{align}

\begin{align}
    \begin{split}
        \mathit M, S_{M} = F_{dec}(E_{answer}, E_{img})
    \end{split}
\end{align}

\subsection{Mask Fusion and Refinement in 3D}
The obtained 2D segmentation mask is eventually needed to be fused in 3D space to obtain the desired 3D segmentation result. We find that the result from directly merging the multi-view segmentation may not be coherent and high-quality due to the accumulated error and lack of comprehensive multi-view 3D information. Therefore, we designed a multi-stage fusion and refinement mechanism to fully exploit the semantic information and viewpoint information to obtain better 3D segmentation results.

First, we use the top-k method to filter the generated masks to reduce errors in 2D segmentation. Specifically, if the area difference between two masks is greater than a certain threshold $T$, we select k=1, indicating that is the mask (the most salient part) we want, we generate a bounding box that fits with the mask; otherwise, we select multiple masks and generate multiple bounding boxes. The filtered top-k masks $S_{M}$, the corresponding confidence scores $S_{M}$, and the face ID is then used as the input to the fusion algorithm. We use the mapping relationship $W_{pf}$ that maps the 2D image mask regions onto the faces of the 3D mesh, resulting in an initial segmented mesh. Note that only the masks within the generated bounding box are involved in the fusion process. 

Following \cite{SATR}, we smooth and refine the segmentation boundaries, reducing noise and errors with Gaussian Geodesic Reweighting. Subsequently, we apply the Visibility Smoothing technique to eliminate discontinuities caused by changes in viewpoints, ensuring that the segmented mesh appears natural and coherent from all angles. Finally, we use a Global Filtering Strategy that filters out the masked regions with low confidence scores.

Specifically, for each 2D mask $M$, we estimate its central face $G_i^j$, where ${i}$ denotes the view and ${j}$ denotes the mask within the view. For the 3D mesh under the current view, we retrieve the vertices of all faces corresponding to the current mask and compute their area-weighted average. This average point is then projected onto point $F$, and the face $F_{i}^j$ containing this projection is identified as the central face for the current view and mask. Subsequently, we calculate the geodesic distance vector $d_i^j\in R^N$ from the central face $G_i^j$ to $F_i^j$ for all faces in $f \in F_i^j$. Here, $N$ represents the number of faces in the mesh for the current mask.

\begin{align}
    f(x) = \left\{
    \begin{aligned}
        &gdist(G_i^j, f), & \text{if } f\in F_i^j \\
        &0, & \text{otherwise}
    \end{aligned}
    \right.
\end{align}
Where $gdist(·,·)$ represents the geodesic length between two faces computed using a heat method \cite{Crane_Weischedel_Wardetzky_2017} on mesh $F$. The geodesic distance between mesh faces measures the path length along the surface from one face to another. 

Next, we fit a Gaussian distribution on the distances and calculate the corresponding probability density values given the geodesic distances between each face and the uppercase face.

\begin{align}
    \begin{split}
        \mathit r_i^j = {\xi[(d;\mu_i^j,(\sigma_i^j)^2)}, d \in d_i^j]
    \end{split}
\end{align}

Where $\mu_i^j$ and $\sigma_i^j$ represent the mean and standard deviation of the distances to $d_i^j$, respectively. Subsequently, we tally the number of times $n$ each face in the mesh is segmented in each view. Finally, we multiply the frequency of each face by the corresponding probability density, and then by the corresponding confidence score $S_M$, to obtain the final confidence for each mesh face. 

However, using only the above method may result in insufficient weighting around the central face $G_i^j$, especially in regions where the average distances between faces are large. To address this issue, we use computes its local neighborhood, where neighbors are determined by mesh connectivity: if two faces share at least one vertex, then face $m$ is considered a neighbor of face $n$. To achieve this, we construct a $q$-rank neighborhood $N_q(n)$ ($q=5$) as follows. For face $m \in F$, if there exists a path on the graph connecting $m$ and $n$ with at most $q$ other vertices along the path, then we include face $n \in F$ in the neighborhood.

Finally, we adopt a global filtering using the calculated threshold. We filter out masked regions with low confidence scores. The threshold is the mean confidence score calculated for every face.

\section{Experiment}
\label{sec:blind}

\subsection{Experimental Setup}
\subsubsection {Dataset and Evaluation Metric:} Since there are no existing zero-shot reasoning 3D segmentation benchmarks, we first evaluated the zero-shot open-vocabulary segmentation performance on the FAUST \cite{FAUST} benchmark (an open-vocabulary 3D segmentation benchmark) proposed in SATR \cite{SATR}. We also validated the effectiveness of our method on reasoning 3D segmentation by our collected in-the-wild data from SketchFab. The FAUST dataset consists of manually annotated registered meshes of human body scans, re-meshed independently for each scan to contain approximately 20K triangular faces. We randomly collected samples from the 3D modeling website SketchFab and asked human volunteers to give implicit segmentation commands. For the evaluation metric, we employ the mean Intersection over Union (mIoU) for semantic segmentation as described in \cite{PartNet} for qualitative evaluation for each semantic category across all test shapes in open-vocabulary 3D segmentation. For the reasoning 3D segmentation, the result is visualized and rated by the user.

\subsubsection {Implementation Details:} We utilized a single NVIDIA A100 GPU for each set of experiments. During the rendering process, we centered the input mesh at the origin and normalized it within a unit sphere. We evenly sample 8 images horizontally around all 360 degrees, maintaining consistency in viewpoints across all experiments. During the rendering process, we used a resolution of 1024$\times$1024 and set a uniform black background color. Multiple reasons (or explanations) will be generated in each view to give a comprehensive understanding for the object, and users can choose one as the desired answer.

\subsubsection{Comparison Experiments for Open-Vocabulary Segmentation}
Since there is no existing reasoning 3D segmentation approach that can be compared, we first compared our method with existing open-vocabulary 3D segmentation models such as SATR \cite{SATR} and 3DHighlighter \cite{3DHighlighter} following the protocol in \cite{SATR} but use the same rendering protocol in our method. As illustrated in Table. \ref{tabe} and Table. \ref{tabf} We show that though not designed for open-vocabulary segmentation tasks and without fine-tuning or specially designed structure, our method achieves competitive performance in the open-vocabulary segmentation benchmark.

    
    
    

\begin{table}[h]
\centering
\caption{Performance on the coarse-grained semantic segmentation on FAUST dataset}\label{tabe}
\begin{tabular}{cccccc}
\hline
Model & Backbone & Arm & Head & Leg & Torso \\

3DHighlighter & CLIP & 28.60 & 14.20 & 14.90 & 8.20 \\

SATR & GLIP & 61.54 & \textbf{76.89} & \textbf{87.4}1 & \textbf{52.32} \\

\textbf{Ours} & \textbf{LISA} & \textbf{64.65} & 72.60 & 83.58 & 50.39 \\
\hline
\end{tabular}
\end{table}

\begin{table}[h]
\centering
\caption{Performance on the fine-grained semantic segmentation on FAUST dataset}\label{tabf}
\begin{tabularx}{\textwidth}{cccccccccccccccccc}
\hline
Model  & Arm & Belly button & Chin & Ear & Elbow & Eye & Foot & Forehead & Hand   \\

3DHighlighter  & 18.39 & 1.99 & 0.46 & 0.72 & 0.08 & 0 & 20.81 & 0.70 & 0.02  \\

SATR  & 24.23 & \textbf{22.00} & \textbf{26.53} & \textbf{34.55} & \textbf{33.67} & \textbf{22.55} & \textbf{75.20} & \textbf{30.35} & \textbf{75.11}  \\

\textbf{Ours}  &  \textbf{26.47} &  1.87 &  3.36 &  10.61 &  18.18 &  2.77 &  71.85 &  6.56 &  43.15  \\
\hline
 & Head & Knee & Leg & Mouth & Neck & Nose & Shoulder & Torso \\
 3DHighlighter & 3.49 & 6.17 & 3.91 & 0.05 & 1.94 & 0.07 & 0.04 & 7.28    \\
 SATR & \textbf{40.31} & \textbf{46.96} & 56.5 & \textbf{20.46} & \textbf{22.01} & \textbf{37.41} & \textbf{24.41} & \textbf{50.52} \\
 \textbf{Ours} &  39.81 &  13.95 &  \textbf{62.23} &  4.12 &  11.88 &  5.5 &  9.6 &  48.78 \\
\hline
\end{tabularx}
\end{table}

\begin{figure*}[h]
\centering
\includegraphics[width=0.8\textwidth]{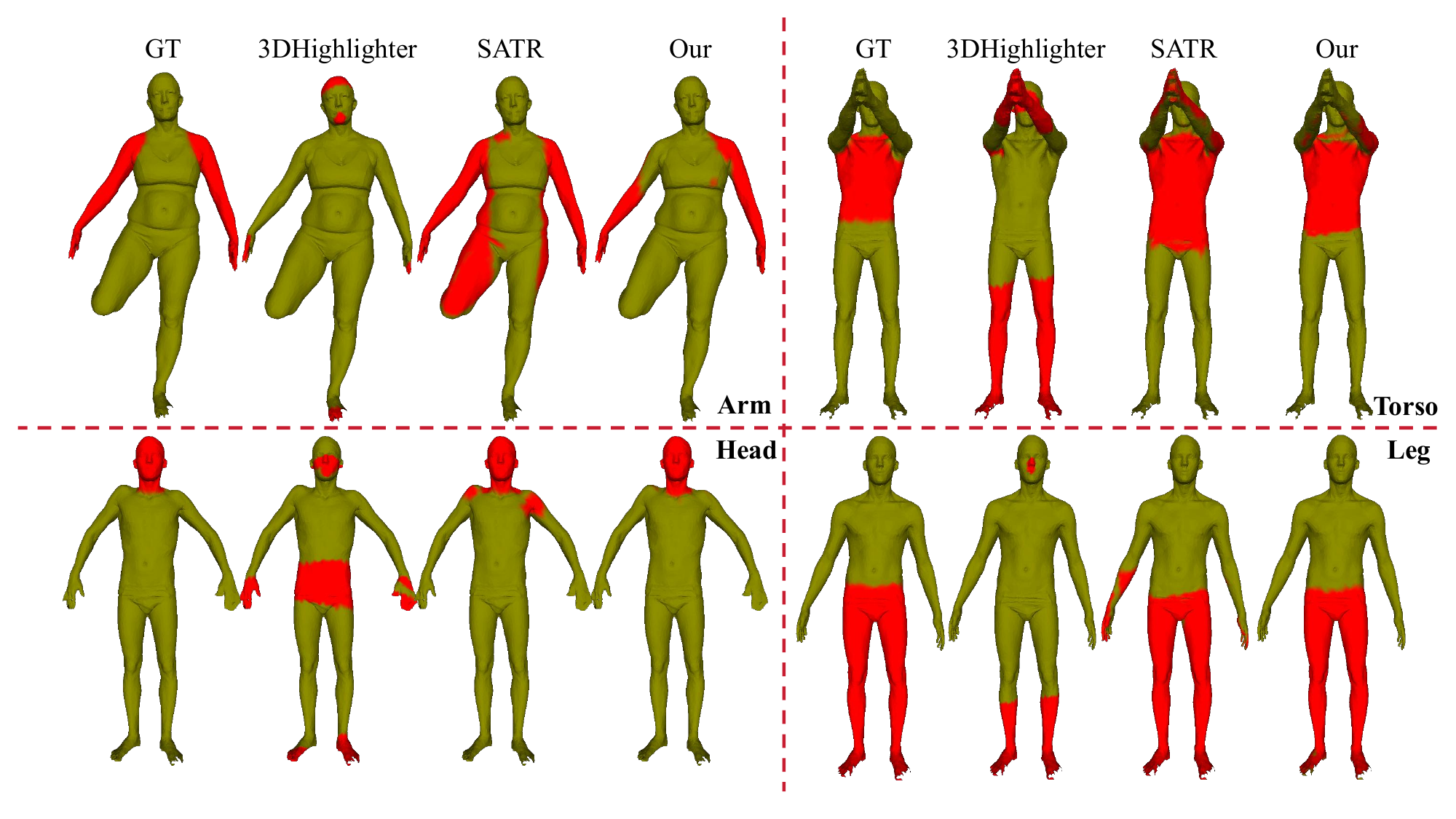}
\caption{Qualitative results and comparison between our method and baseline method in FAUST benchmark. The segmented regions are shown in red.} \label{exp2}
\label{sketchin}
\end{figure*}

\subsection{Performance in Reasoning 3D Segmentation}
\begin{figure*}[h]
\centering
\includegraphics[width=\textwidth]{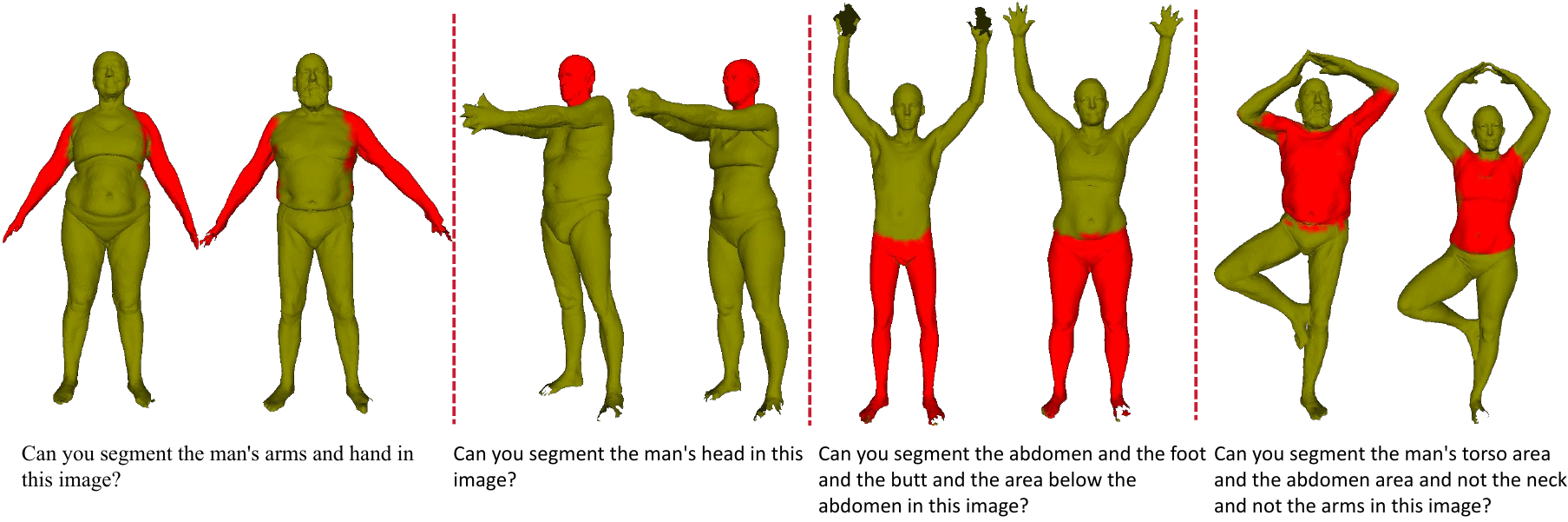}
\caption{A natural language command can make the model segment the desired regions. The segmented regions are shown in red.} \label{exp5}
\label{sketchin}
\end{figure*}

A better property that our method has compared to existing open-vocabulary segmentation is that our method can use natural language as the input information. The LLM parses the natural language and gives the segmentation result directly, which enables a more natural and convenient computer-human interaction experience. An example is shown in Fig. \ref{exp5}. The models are from the FAUST dataset.

In the open-vocabulary segmentation, only explicit segmentation command is given, in which Reasoning3D's potential has not been fully exploited. We randomly collect 3D models from the 3D modeling website SketchFab perform the assessment with these in-the-wild 3D models and let human volunteers give ``implicit" segmentation commands. Figure. \ref{exp3} and fig. \ref{fig1} shows some examples. The examples show that Reasoning3D has the capabilities to offer in-depth reasoning, 3D understanding, part segmentation, and conversational abilities. The model can output the segmentation masks and the explanation as we desire. 

To better allow users to interact with our system, we also designed a User Interface (UI) so that users can input arbitrary 3D models and their desired prompt to segment the desired region. (Fig. \ref{sketchin}) This UI will also be open-sourced. 

\begin{figure*}
\centering
\includegraphics[width=\textwidth]{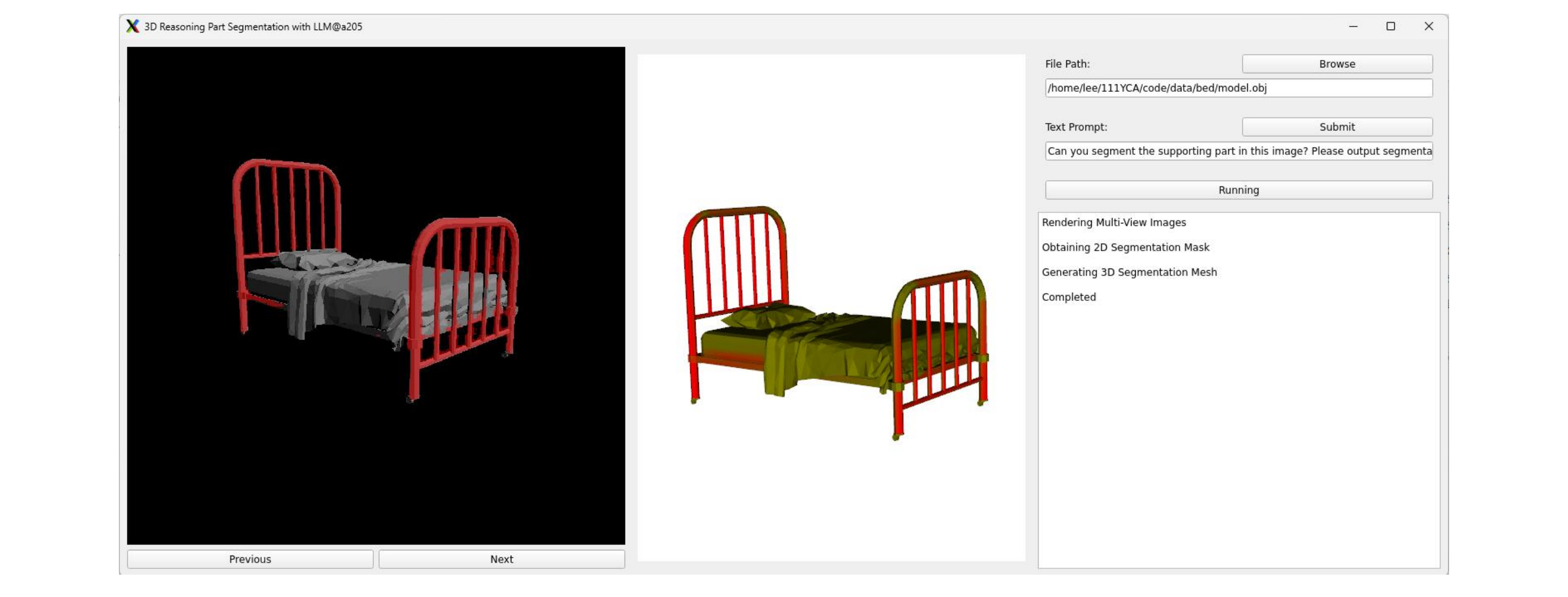}
\caption{We offer a user-friendly interface designed for performance assessment, facilitating the easy upload of 3D models and prompts by users. It enables swift acquisition of 3D segmentation outcomes. This tailored software is available as open-source.}\label{exp3}
\label{sketchin}
\end{figure*}

\begin{figure*}
\centering
\includegraphics[width=\textwidth]{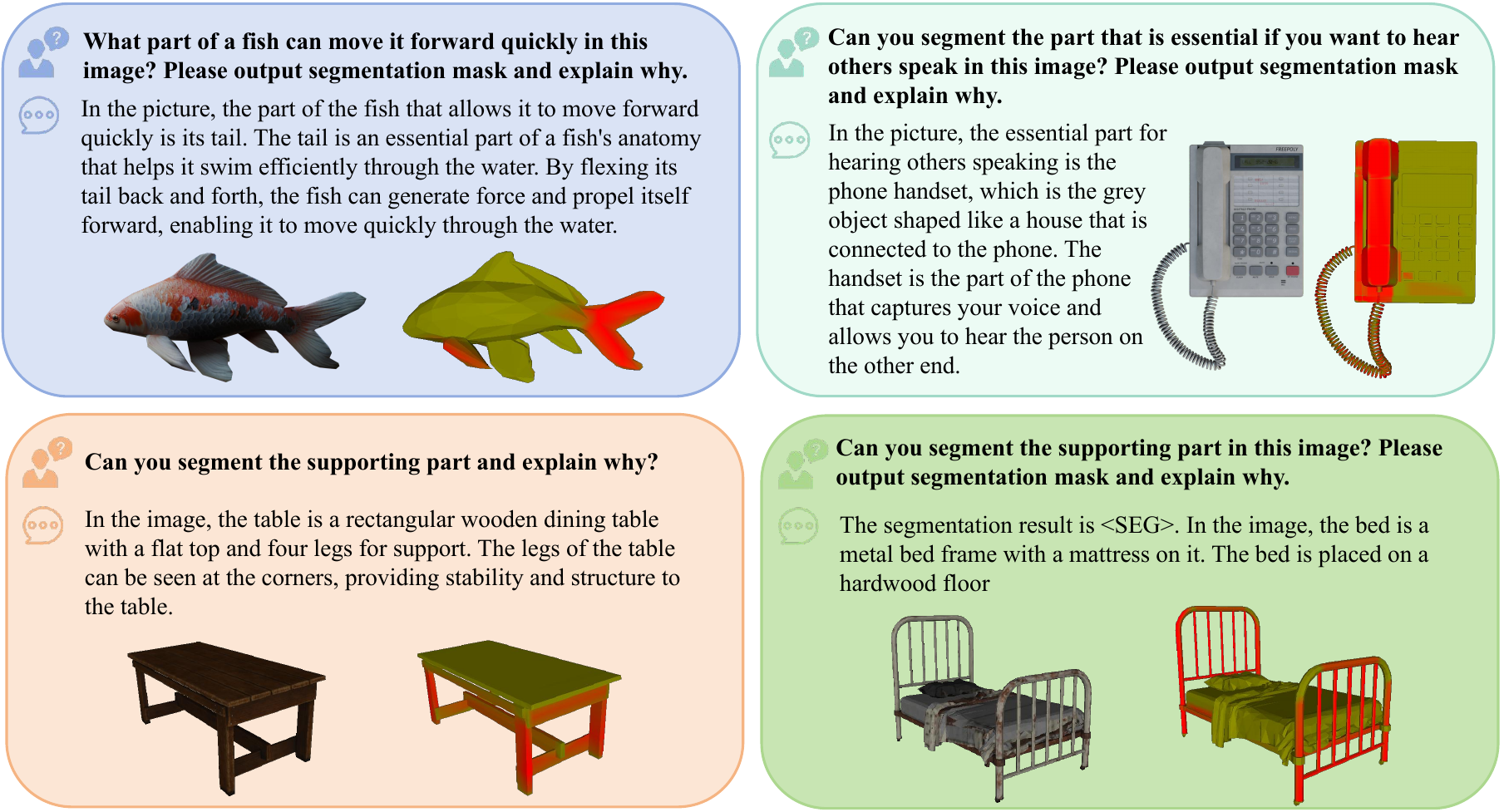}
\caption{This figure shows Reasoning3D's ability to segment 3D object parts (in a fine-grained manner) from in-the-wild samples, including real-world scanned data (samples are randomly collected from SketchFab). These examples highlight Reasoning3D's advanced capabilities in in-depth reasoning, comprehensive 3D understanding, precise part segmentation, and robust conversational abilities. The original mesh and the segmentation result are visualized, and the segmented region is highlighted in Red. }\label{exp3}
\label{sketchin}
\end{figure*}

\section{Discussion and Limitations}
This research represents preliminary findings in the task of reasoning 3D segmentation, and several areas require further exploration and validation. One major aspect is the need for comprehensive benchmarking to rigorously evaluate our method's performance. Additionally, conducting user studies will provide valuable insights into the practical applicability and usability of our approach.

Our findings indicate that view information plays a critical role in the performance of 3D segmentation tasks. Optimizing view selection to align with the pre-trained vision encoder could significantly enhance outcomes. This suggests that a strategic approach to view selection is essential for leveraging the full potential of the pre-trained models.

The flexibility of our method is noteworthy, as the LVLM can perform zero-shot inference without the need for additional training. While fine-tuning with data could potentially improve performance, we observed that fine-tuning with a very small dataset might negatively impact the network's generalization ability, sometimes resulting in poorer performance compared to fine-tuning. It is also worth noting that our multi-view 2D segmentation and 3D projection method can be applied to scenes, which will be beneficial for more real-world applications.

To foster further advancements and collaborative innovation in 3D reasoning and segmentation, we are releasing our code. We encourage the community to build upon our work and develop improved methods.

\section{Conclusion}
This paper introduces a new task: Zero-Shot 3D Reasoning Segmentation for part searching and localization within objects. This new approach moves beyond the constraints of traditional category-specific 3D semantic segmentation, 3D instance segmentation, and open-vocabulary 3D segmentation. We have developed Reasoning3D, a simple yet effective baseline method that can understand and perform complex commands to segment specific parts of 3D meshes with contextual understanding and reasoned outputs for interactive segmentation.

Reasoning3D leverages pre-trained 2D segmentation networks in conjunction with Large Language Models (LLMs) to interpret user queries in a zero-shot manner. Previous studies have shown that extensive pre-training equips foundational models with a broad understanding of the world, enabling them to process complex commands. Our method harnesses this capability, allowing for effective 3D segmentation with limited 3D datasets, making it a resource-efficient solution.

Our experiments demonstrate that Reasoning3D is generalizable and capable of accurately localizing and identifying parts of 3D objects based on implicit textual queries. This includes both articulated 3D objects and real-world scanned data. Additionally, our method can produce natural language explanations for the segmented 3D models and their components. The training-free nature of our approach facilitates rapid deployment and provides a robust baseline for future research in part-level 3D object understanding. This has potential applications across various domains, such as robotics, object manipulation, part assembly, autonomous driving, augmented and virtual reality (AR/VR), and medical fields.

We are releasing the code, model weights, deployment guide, and evaluation protocol to encourage further innovation and collaboration. These resources are available at: http://tianrun-chen.github.io/Reason3D/.

%
%
\bibliographystyle{splncs04}
\bibliography{main}
\end{document}